\documentclass{article} %
\usepackage[T1]{fontenc}
\usepackage[preprint]{colm2025_conference}

\usepackage{microtype}
\usepackage{hyperref}
\usepackage{url}
\usepackage{booktabs}

\usepackage{algorithm}
\usepackage{algorithmicx}
\usepackage{algpseudocode}
\usepackage{amsmath}
\usepackage{enumitem}
\usepackage{graphicx}
\usepackage{adjustbox}

\definecolor{Magenta}{rgb}{0.8, 0.1, 0.6}

\usepackage{lipsum}
\usepackage{wrapfig}
\usepackage{subcaption}

\usepackage{lineno}

\definecolor{darkblue}{rgb}{0, 0, 0.5}
\hypersetup{colorlinks=true, citecolor=darkblue, linkcolor=darkblue, urlcolor=darkblue}

\title{Routing to the Right Expertise: \\ A Trustworthy Judge for Instruction-based Image Editing}

\author{Chenxi Sun, Hongzhi Zhang, Qi Wang \thanks{Corresponding Author}~,~  Fuzheng Zhang\\
Kuaishou Technology\\
\texttt{sunchenxi@kuaishou.com} \\
}

\begin{document}

\ifcolmsubmission
\linenumbers
\fi

\maketitle

\begin{abstract}
 Instruction-based Image Editing (IIE) models have made significantly improvement due to the progress of multimodal large language models (MLLMs) and diffusion models, which can understand and reason about complex editing instructions. In addition to advancing current IIE models, accurately evaluating their output has become increasingly critical and challenging. Current IIE evaluation methods and their evaluation procedures often fall short of aligning with human judgment and often lack explainability. To address these limitations, we propose \textbf{JU}dgement through \textbf{R}outing of \textbf{E}xpertise (\textbf{JURE}). Each expert in JURE is a pre-selected model assumed to be equipped with an atomic expertise that can provide useful feedback to judge output, and the router dynamically routes the evaluation task of a given instruction and its output to appropriate experts, aggregating their feedback into a final judge. JURE is trustworthy in two aspects. First, it can effortlessly provide explanations about its judge by examining the routed experts and their feedback. Second,
 experimental results demonstrate that JURE is reliable by achieving superior alignment with human judgments, setting a new standard for automated IIE evaluation. Moreover, JURE’s flexible design is future-proof - modular experts can be seamlessly replaced or expanded to accommodate advancements in IIE, maintaining consistently high evaluation quality. Our evaluation data and results are available at \url{https://github.com/Cyyyyyrus/JURE.git}.

\end{abstract}

\section{Introduction}

Recent advancements in multimodal large language models (MLLMs) and diffusion models have significantly expanded their capabilities, enabling sophisticated multimodal reasoning and generative tasks. Among these, Instruction-based Image Editing (IIE)~\citep{DBLP:journals/corr/abs-2404-09990, DBLP:conf/cvpr/BrooksHE23, DBLP:journals/corr/abs-2408-14180} has emerged as a critical area, where models modify images based on natural language instructions.
Approaches such as Emu Edit~\citep{DBLP:conf/cvpr/SheyninPSKZAPT24} and SeedEdit~\citep{shi2024seededitalignimageregeneration} further demonstrated that leveraging high-quality instructional paired data
can significantly improve the editing quality. Despite these advances, evaluating the quality of edited images remains challenging.
Human evaluation, while effective, is costly, time-consuming, and prone to inconsistency, underscoring the need for a reliable automated evaluation framework.

Evaluation methods for IIE initially relied on conventional metrics by comparing the edited image with a reference. 
For example, CLIP-based methods~\citep{radford2021learning}, which, despite their simplicity and objectivity, fail to capture nuanced semantic alignment. Peak Signal-to-Noise Ratio (PSNR)~\citep{korhonen2012peak} measures pixel-level similarity, but in style editing tasks, substantial pixel variations—such as changes in texture or color—may still preserve the intended artistic style, despite resulting in a low PSNR score. To address these limitations, recent methods tailored for IIE~\citep{kawar2023imagic, basu2023editval, DBLP:conf/nips/MaJYLWZZSJ24} focus on evaluating models' adherence to the editing instruction. However, these evaluation methods have limitations such as limited dataset diversity (as seen in~\citet{kawar2023imagic}), narrow applicability (e.g., ~\citet{wang2023imagen} for mask-guided editing only) or restricted flexibility for complex editing instructions (e.g., \citet{DBLP:conf/cvpr/HuangXWYCG00HZS24} and \citet{DBLP:journals/corr/abs-2408-14180}). Most importantly, none of them achieves human-level judgment quality.

\begin{figure*}[t]
    \centering
    \begin{adjustbox}{width=1.02\linewidth,center}
        \includegraphics{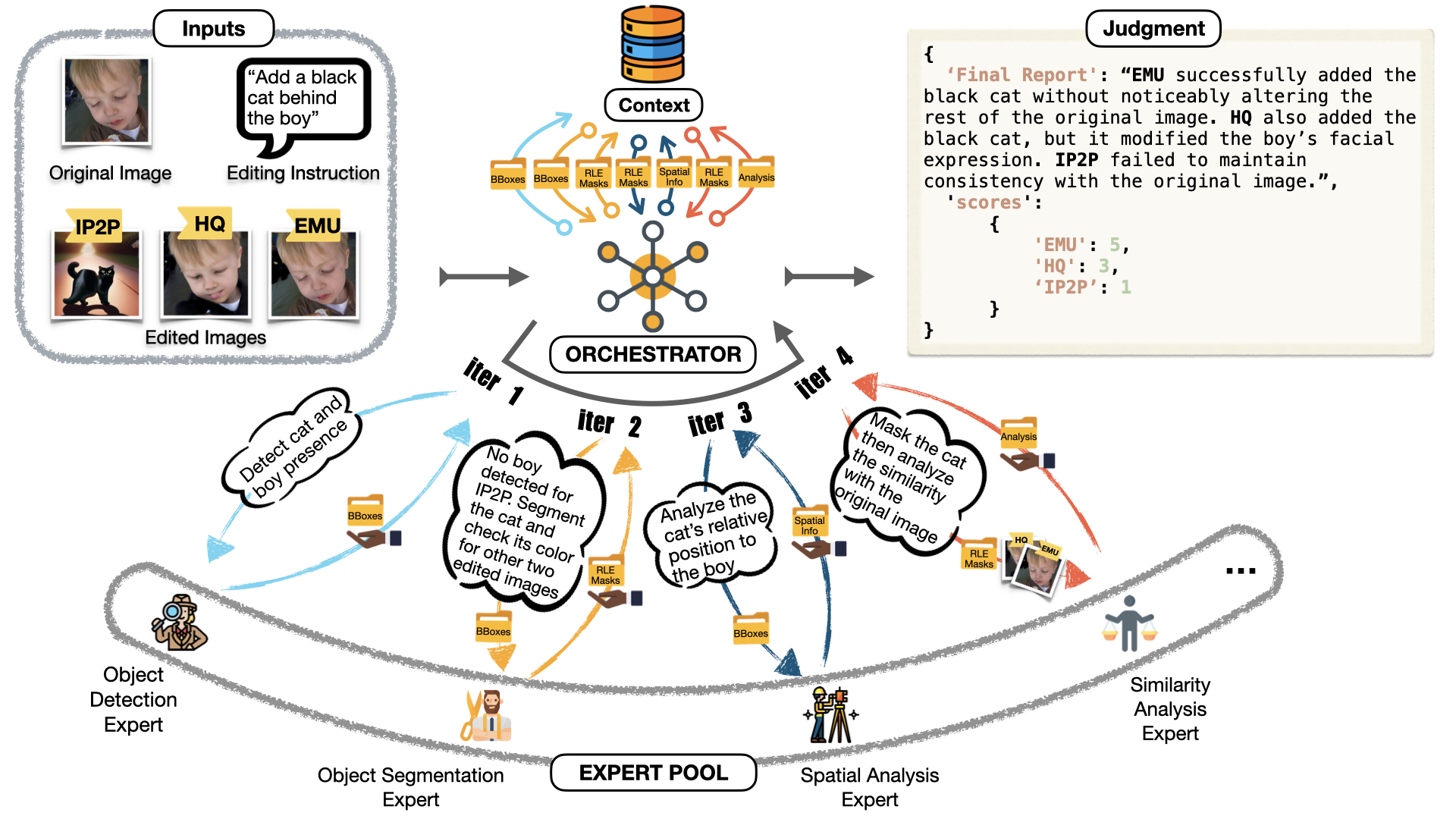}
    \end{adjustbox}
    \caption{Conceptual architecture of the JURE framework.
    JURE takes the original image, editing instruction ``Add a black cat behind the boy'', edited images generated by different IIE models (IP2P, HQ Edit, EMU Edit) as the initial input. In this example, JURE iteratively reasons about to verify \emph{(i)} object presence (whether a cat was added), \emph{(ii)} attribute accuracy (if added, whether it is black), \emph{(iii)} spatial correctness (whether it appears behind the boy), and \emph{(iv)} visual integrity (ensuring no unwanted edits or artifacts appear elsewhere). During each iteration (highlighted by different colors), the Orchestrator routes to the right expert, incrementally stores their output to the \emph{Context Dictionary} for future reference, and dynamically decides its action for the next iteration. For instance, after detecting that IP2P failed to retain the boy in the first iteration, subsequent spatial analysis involving the boy and the cat is performed only for the HQ and EMU edits. Finally, it aggregates all expert responses to produce a final judgment.
    }
    \label{fig:jure-architecture}
\end{figure*}

Another significant field that has benefited from recent advancements in LLMs and MLLMs is automated judgment, which involves leveraging these models as powerful evaluators. Proprietary models like GPT-4o have demonstrated exceptional human-aligned judgment, across evaluations of conversational AI systems~\citep{DBLP:conf/nips/ZhengC00WZL0LXZ23}, creative writing~\citep{app15062971}, visual question-answering~\citep{DBLP:journals/corr/abs-2311-01361}, among others. Some other studies have also explored finetuning current LLMs/MLLMs into critic models
~\citep{DBLP:conf/iclr/WangYYZYW0J000024,DBLP:journals/corr/abs-2410-02712}. 
However, finetuning a critic model specialized for IIE may suffer from a lack of scarcity domain-specific finetuning data and heavy post-training cost. Moreover, our experiment results reveal the poor judgment quality of strong general MLLMs like OpenAI o1 when applied to IIE evaluation.

Motivated by these challenges, we propose the \textbf{JU}dgment via \textbf{R}outing of \textbf{E}xpertise (\textbf{JURE}) framework for IIE. We argue that evaluating IIE outputs can be decomposed into multiple atomic evaluation dimensions, each assessed by specialized models with distinct, minimally overlapping expertise. Specifically, JURE employs an expertise-aware \textbf{Orchestrator}, instantiated with a powerful MLLM, to dynamically route evaluation tasks to the most suitable experts within a \textbf{Expert Pool}. This structured routing mechanism allows JURE to serve as an explainable and general evaluation framework for IIE and sidestep the limitations of monolithic evaluators.
Experimental comparisons with strong evaluators and human annotators demonstrate that JURE achieves superior alignment with human judgment, rivaling human inter-annotator agreement. Further analysis of the Orchestrator’s routing decisions confirms that \emph{(i)} relevant experts are precisely invoked based on task requirements, and \emph{(ii)} expert feedback meaningfully contributes to accurate and explainable final evaluations, highlighting the effectiveness and interpretability of our framework.

Our main contributions are summarized as follows:
\begin{itemize}[noitemsep, leftmargin=*]
    \item \textbf{A reliable evaluation framework for IIE.}
    We propose JURE, a novel framework for Instruction-based Image Editing (IIE) that uses expertise routing to achieve superior alignment with human judgment.

    \item \textbf{Expertise Routing for explainable judgment.}  
    Experiments show JURE effectively invokes relevant experts based on task-specific evaluation needs, demonstrating the feasibility and benefits of expertise-driven routing.

    \item \textbf{A Modular and Explainable Framework.}  
    JURE provides interpretable, dimension-specific feedback and adapts seamlessly to evolving evaluation criteria with minimal overhead.
    
\end{itemize}

\section{Related Work}

\subsection{IIE Evaluations}
The evaluation methods for IIE start with the exploitation of conventional metrics.
For example, Peak Signal-to-Noise Ratio (PSNR)~\citep{korhonen2012peak} and Similarity Index Measure (SSIM)~\citep{wang2004image} objectively assess pixel fidelity and structural consistency, but they fail to capture perceptual and semantic correctness, particularly for complex edits. Learned Perceptual Image Patch Similarity (LPIPS)~\citep{welling2011bayesian} improves alignment with human perception by leveraging deep neural embeddings but remains limited to reference-based comparisons. CLIP-based metrics~\citep{radford2021learning} provide a measure of semantic alignment but insufficiently account for unintended visual alterations and perceptual realism.

Some recent works tailored specifically for IIE have also been proposed.
They usually involve specialized automated methods designed to assess how effectively edits align with high-level user instructions and maintain visual coherence. EditBench~\citep{wang2023imagen} only works for mask-guided editing by measuring mask-region accuracy and background preservation, limiting their applicability to general editing methods. Reason-Edit~\citep{DBLP:conf/cvpr/HuangXWYCG00HZS24} assesses complex, multi-step edits by evaluating both semantic alignment and logical consistency in executing sequential transformations. Reason-Edit incorporates structured reasoning metrics to verify whether edits follow the intended multi-step modifications while preserving coherence in unrelated image regions. However, its reliance on predefined reasoning structures limit its flexibility in evaluating complex out-of-domain instruction. I2EBench~\citep{DBLP:journals/corr/abs-2408-14180} introduces sixteen evaluation dimensions with different automated metrics but assumes each instruction fits neatly into a single predefined category, limiting its ability to assess complex, multi-faceted edits. 
Despite these advancements, no existing IIE evaluation matches human-level judgments. Developing an automated evaluator closely correlated with human perception remains a crucial research direction.

\subsection{(M)LLM-as-Judge}

From a broader perspective of automated evaluation, researchers have explored using strong proprietary models~\citep{openai2024gpt4ocard, geminiteam2024gemini15unlockingmultimodal} as ``judges''~\citep{DBLP:conf/nips/ZhengC00WZL0LXZ23, app15062971}.
~\citet{DBLP:conf/nips/ZhengC00WZL0LXZ23} validates GPT-4's strong evaluation on MT-Bench and Chatbot Arena can match human preferences. ~\citet{DBLP:journals/corr/abs-2212-08073} incorporates AI feedbacks on alignment tasks, while works for evaluating visual question-answering~\citet{DBLP:journals/corr/abs-2311-01361} further extend its application to multi-modal evaluation.
Although these methods yield reliable judgments, they lack the customizability required for domain-specific tasks due to their close nature. To address these limitations, recent efforts in the open-source community have explored training publicly accessible critic models and enhancing them with better explainability~\citep{DBLP:conf/iclr/WangYYZYW0J000024, DBLP:journals/corr/abs-2308-04592, DBLP:journals/corr/abs-2410-16256, DBLP:conf/iclr/LiSYF0024, DBLP:conf/emnlp/KimSLLSWNL0S24}, with PandaLM~\citep{DBLP:conf/iclr/WangYYZYW0J000024} as an early endeavor towards explainable judgment and LLaVA-Critic~\citep{DBLP:journals/corr/abs-2410-02712} as a milestone towards explainable multi-modal judgments. However, while training an ``all-singing, all-dancing'' large judge model seems promising, the No Free Lunch (NFL) Theorem~\citep{DBLP:journals/tec/DolpertM97} highlights the inherent capacity trade-offs they face. A single model with fixed parameter space must balance competing evaluation objectives, leading to comprised performance. Fine-tuning with targeted-domain data is a popular method but the in-domain data is often scarce or unavailable. Even when fine-tuning is feasible, it still requires resource-intensive retraining or complex continual training strategies to ensure alignment with the evolving capabilities of the generative models being evaluated. Beyond monolithic judge models, recent works~\citep{DBLP:journals/corr/abs-2410-10934} have explored agent-based approaches that combine multiple roles using an LLM or MLLM with different prompts. However, these agent systems are trivial and limited, as they rely on the same underlying capabilities across roles, offering no genuine specialization or diversity in expertise.

\section{Instruction-based Image Editing (IIE)}
\label{sec:iie}

\begin{wraptable}{r}{0.5\textwidth}
    \vspace{-13pt}
    \centering
    \caption{Representative sub-tasks in IIE along with Emu testset distribution.}
    \label{tab:iie-subtasks}
    \begin{tabular}{l r}
        \toprule
        \textbf{Sub-task} & \textbf{\#Instances (\%)} \\
        \midrule
        Object Addition & 550 (28.59\%) \\
        Object Replacement & 204 (10.60\%) \\
        Object Movement & 13 (0.68\%) \\
        Object Removal & 285 (14.81\%) \\
        Background Change & 391 (20.32\%) \\
        Attribute Change & 322 (16.74\%) \\
        Style Change & 241 (12.53\%) \\
        Size/Shape Modification & 6 (0.31\%) \\
        Perspective Editing & 4 (0.21\%) \\
        \bottomrule
    \end{tabular}
    \vspace{-5pt}
\end{wraptable}

Instruction-based Image Editing (IIE) is an emerging paradigm that allows users to modify images using textual instructions. Unlike traditional image editing, which requires manual manipulation of pixels, layers, or masks, IIE enables users to describe desired modifications in natural language (e.g., ``make the sky look like a sunset''), and an AI model autonomously applies the edits. 
The rapid advancement of IIE~\citep{DBLP:conf/cvpr/SheyninPSKZAPT24, DBLP:journals/corr/abs-2404-09990, DBLP:conf/cvpr/BrooksHE23, DBLP:journals/corr/abs-2408-14180} has been driven by breakthroughs in MLLMs and diffusion-based image generation, which can be trained on paired data consisting of editing instructions and corresponding edited image to unlock this ability.

While different works propose varying categorizations of IIE sub-tasks, we  adopt a structured taxonomy inspired by~\citet{DBLP:conf/cvpr/SheyninPSKZAPT24},
summarized in Table~\ref{tab:iie-subtasks} with detailed examples provided in Table~\ref{appendix:tab:iie-subtasks} (\S\ref{appendix:iie-subtasks}).

\section{The JURE Framework}
\label{sec:jure-framework}

In this section, we present the design of our Judgment through Routing of Expertise (JURE) framework for IIE. We begin with a high-level overview (\S\ref{sec:jure-overview}), then describe the modular \emph{Expert Pool} (\S\ref{sec:expert-pool}) and the expertise-aware \emph{Orchestrator} responsible for routing of expertise and aggregation of results (\S\ref{sec:orchestrator}). Finally, we provide implementation details regarding microservices and infrastructure (\S\ref{sec:infrastructure}).

\subsection{Overview of JURE}
\label{sec:jure-overview}

The key idea behind JURE is to treat automated evaluation as a decomposable process, where evaluation objectives are broken down into specialized sub-tasks requiring atomic expertise, rather than relying on a monolithic scoring model. Given an editing instruction $T$ and an original image $I$, JURE dynamically decomposes the evaluation into a set of atomic evaluation dimensions ${c_1, \dots, c_k}$ based on the specific instruction. Each dimension $c_i$ is assessed by one or more specialized expert models $m_i$. While atomic evaluation dimensions and experts are not one-to-one, the Expert Pool collectively covers all required expertise.

Figure~\ref{fig:jure-architecture} offers an example of JURE’s architecture and workflow.
As depicted, there are two central components:

\begin{itemize}[noitemsep, leftmargin=*]
  \item \textbf{Expert Pool.} 
  A set of specialized models (Experts), each providing a particular evaluation expertise. 
  The implementation of experts are flexible. It can output a score or a brief structured comment, even though, their focus on \emph{one atomic expertise} enables JURE to identify precisely which sub-task is done well or needs improvement.

  \item \textbf{Orchestrator.} 
  A capable MLLM that (1) \emph{routes} dynamically to necessary experts based on the user’s editing instruction, (2) \emph{aggregates} the experts’ responses (e.g., scalar scores, textual feedback, or event extracted features), and (3) \emph{synthesizes} a final verdict with scores and natural-language explanations.
\end{itemize}

By decomposing complex evaluation problems into specialized checks, JURE enjoys several advantages. It is \textbf{modular} and \textbf{extensible}: individual experts can be added, updated, or removed without re-engineering an entire end-to-end evaluator to evolve with IIE models' advancements. 
Besides, it is \textbf{interpretable}: the atomic nature of each expert’s ability clarifies which components of a generated output excel or fail, in contrast to returning a single opaque quality score.

\begin{wraptable}{r}{0.5\textwidth}
    \vspace{-13pt}
    \centering
    \caption{Expert Pool for IIE Evaluation.}
    \label{tab:iie-experts}
    \begin{tabular}{l p{0.47\linewidth}}
        \toprule
        \textbf{Expert} & \textbf{Implementation} \\
        \midrule
        Object Detection  & DINO-X \\
        Object Segmentation  & SAM-2 \\
        Depth Analysis  & DepthAnything-V2 \\
        OCR  & GPT-4o \\
        Similarity Analysis  & GPT-4o \\
        Style Analysis  & GPT-4o \\
        Chroma and Pattern  & GPT-4o \\
        \bottomrule
    \end{tabular}
    \vspace{-16pt}
\end{wraptable}

\subsection{The Expert Pool}
\label{sec:expert-pool}

JURE is grounded in the notion of \emph{atomic expertise}, 
where each expert model is responsible for a narrowly scoped evaluation skill (e.g., object detection, style analysis). 
This principle motivates us to construct an expert pool that covers the various dimensions of IIE evaluation. To ensure ease of reproduction and rapid verification, we prioritize the use of off-the-shelf models (Open-source or API-based). Table~\ref{tab:iie-experts} lists the experts and their mapped expertise.

\paragraph{Atomic Expertise.}
An atomic expertise is a well-defined capability that maps an input to a score or structured feedback. They all adhere to the principle of \textit{singularity of purpose}. For instance, in IIE evaluation, separate experts might provide object addition, depth analysis, or style analysis expertise. This modular approach enhances explainability by allowing each expert to provide focused, interpretable feedback, making it easier to understand and diagnose specific aspects of the evaluation.

\paragraph{Implementations.}
The design of the expert pool follows a coverage-oriented principle, ensuring that the collective capabilities of the experts span all relevant evaluation dimensions. Rather than assigning a single expert to each sub-task in a strict one-to-one manner, one aspect of evaluation can require multiple different experts and vice versa. While the granularity of expert specialization remains flexible, constructing an effective expert pool does require domain knowledge to ensure that the selected models provide comprehensive and reliable assessments across all targeted dimensions.

The actual implementation of each Expert is left to the system designer. Possible strategies include: \emph{(i)} Open-Source models, such as CLIP for semantic matching, DINO for object detection, or specialized classifiers for domain-specific tasks. \emph{(ii)} Custom-trained models, fine-tuned on task-specific datasets to ensure better alignment with niche evaluation needs. \emph{(iii)} Rule-based or hybrid approaches for cases where a neural model is unnecessary or infeasible (e.g., checking edited image's RGB range).

To ensure seamless integration with the Orchestrator, each Expert must provide a structured self-description covering three key components. The first is a clear and informative articulation of its \textbf{primary expertise} outlining its specific evaluation capability,
and any key considerations in its invocation.
Equally important is the \textbf{input schema}, which defines the function signature, including the function name, required arguments, and data types. This standardization allows the Orchestrator to invoke it correctly without additional adaptation.
Lastly, the \textbf{output schema} specifies the format and the interpretation methods of the returned results. Outputs may take the form of numerical scores (e.g. average depth from DepthAnything-V2), textual explanations (e.g. OCR results from GPT-4o) or even structured metadata (e.g. RLE masks from SAM-2) that can be fed to other experts. 
Based on our development experience, we find that standardizing input-output schemas across experts significantly improves interoperability and reduces errors in code execution. For example, standardizing the RLE mask format used by SAM-2 and DepthAnything-V2 eliminate execution errors when the Orchestrator chained them together.

\paragraph{Adaptability.}
Since atomic abilities are inherently composable, JURE can evolve alongside advancing IIE models. New evaluation dimensions can be seamlessly integrated by adding specialized experts, while outdated or underperforming ones can be swapped out with minimal overhead. This modularity ensures JURE remains an future-proof framework, consistently delivering a strong judgment signal and adapting to emerging evaluation needs.

\subsection{The Orchestrator}
\label{sec:orchestrator}

While the Expert Pool collectively provides the foundation for multi-dimensional evaluation expertise, the expertise-aware Orchestrator is the core decision-making component.

\paragraph{Dynamic Expert Invocation.}
As shown in Figure~\ref{fig:jure-architecture}, unlike static evaluation pipelines, the Orchestrator dynamically adapts its evaluation strategy based on the given editing instruction and intermediate results from previous expert invocations.
This process ensures that computational resources are used efficiently while the evaluation remains precise, as only the relevant dimensions of evaluation are assessed. With that being said, a qualified Orchestrator should demonstrate superb reasoning ability. Thus, we take advantage of the most capable reasoning model, \textbf{OpenAI o1}~\citep{openai2024openaio1card}, as our Orchestrator.
Please refer to \S\ref{sec:para:workflow} for a detailed algorithmic workflow.

\paragraph{Implementations and Execution.}
At initialization, the Orchestrator receives a structured prompt containing the editing instruction, the original and edited images, and a description of the available experts, including their expertise, input-output schema, and potential usage scenarios. This structured input allows the Orchestrator to understand each expert’s expertise and determine how they can contribute to the evaluation process.

Upon receiving the evaluation request, the Orchestrator analyzes the given instance and determines which experts to call. In each iteration, it invokes experts, analyses their feedback and decide its action for the next iteration until it signals completeness. 
This iterative loop ensures that the evaluation remains contextually aware and adapts to emerging issues rather than rigidly following a pre-defined fixed sequence.

Beyond invoking external experts, when additional image transformations or auxiliary operations are needed, the Orchestrator can also execute codes directly.
To enhance efficiency and robustness, we also provide several pre-built utility functions like image cropping based on RLE masks or bounding boxes. 

Once the evaluation process is complete, the Orchestrator synthesizes the collected information into a structured final report with scores and explanation. The overall execution logic is further illustrated in \S\ref{sec:para:workflow} and formally outlined in Algorithm~\ref{alg:jure-main} in \S\ref{appendix:algorithm}.

\subsection{Infrastructure and Implementation}
\label{sec:infrastructure}

While the Orchestrator and Expert Pool define the conceptual backbone of JURE for IIE, the actual implementation must accommodate flexible scaling, inter-process communication, and a shared context for storing intermediate results. In our design, each Expert is encapsulated as a microservice, and the Orchestrator maintains a \emph{context dictionary} (or state) to coordinate these distributed components.

\paragraph{Microservice Architecture.}
Each expert is deployed behind a lightweight API (e.g., using \texttt{FastAPI}), exposing a consistent interface corresponding to its self-description. For instance, an expert might define an endpoint \texttt{/objectDetectionExpert} that accepts a JSON payload with an image URL and returns a list of detected objects with bounding boxes and confidence scores. By containerizing each expert, the system remains highly modular: updates to one expert (e.g., swapping in a more accurate object detector) do not affect others.

\paragraph{Context Dictionary.}
The Orchestrator maintains a shared \emph{context dictionary} to store metadata, intermediate outputs, and any auxiliary variables generated over the course of evaluation. After each expert is invoked, the Orchestrator is responsible for deciding whether to append the returned information to this dictionary for later consuming. 

\paragraph{Execution Workflow.}
\label{sec:para:workflow}
The main evaluation loop is guided by the orchestrator’s reasoning process, which is abstracted as \textsc{OrchestratorReasoning} in the pseudo-codes in \S\ref{appendix:algorithm}. At each step, the orchestrator analyzes the current state of \texttt{Context} to identify any remaining evaluation gaps. If further assessment is necessary, it selects an appropriate expert constructs corresponding input arguments. 

Each selected expert is queried through \textsc{InvokeExpert}, and returns a structured feedback. The Orchestrator updates \texttt{Context} with these results, allowing subsequent iterations to incorporate new insights into the evaluation process. 
Figure~\ref{fig:jure-architecture} offers a detailed execution example.
Once the Orchestrator deems the evaluation complete, it terminates execution and consolidates the collected findings into a structured final report with scores.

\section{Experiments}
\label{sec:experiments}

\subsection{Dataset}
\label{sec:dataset}

We use the Emu Edit test set~\citep{DBLP:conf/cvpr/SheyninPSKZAPT24}.
To ensure that each instruction corresponds to a well-defined editing sub-task, we use GPT-4o~\citep{openai2024gpt4ocard} to classify each instance into one of nine IIE sub-task categories in Table~\ref{tab:iie-subtasks}. Instances that could not be confidently categorized were discarded.
From an initial set of 2,022 instances, we retain a total of 1,924 samples after filtering, from which we randomly sampled 120 samples for human annotating.

\subsection{IIE Models}
\label{sec:iie-models}

We evaluate three instruction-based image editing models: Emu Edit~\citep{DBLP:conf/cvpr/SheyninPSKZAPT24}, HQ Edit~\citep{DBLP:journals/corr/abs-2404-09990}, and InstructPix2Pix (IP2P)~\citep{DBLP:conf/cvpr/BrooksHE23}. Each model processes the same set of instructions and source images, generating an edited image for evaluation.

\subsection{Human Annotation}
\label{sec:human-annotation}

To establish a reliable human evaluation baseline, we conducted a comparative annotation involving 9 annotators. Annotators were presented with the outputs from all three IIE models. Each annotator rated the outputs of all three models using a 1–5 Likert scale. This method reduces individual bias, leading to more calibrated and discriminative scoring results.

\subsection{Evaluation Metrics}
\label{sec:metrics}

To assess alignment between JURE and human judgments, we employ the Cohen's Kappa Correlation ($\kappa$). The Cohen’s Kappa is a robust statistic for measuring inter-rater reliability.
Since our experiments involve scoring on an ordinal scale, we specifically utilize a quadratically weighted version. The weighting further penalizes disagreements proportionally to their severity.
Formally, the linearly weighted Cohen's Kappa correlation is defined as:

\begin{equation}
\kappa = 1 - \frac{\sum_{i,j} w_{ij}O_{ij}}{\sum_{i,j} w_{ij}E_{ij}}, \quad
w_{ij} = 1 - \frac{(i - j)^2}{(k - 1)^2}
\end{equation}

where \( O_{ij} \) represents the observed frequency of cases assigned to category \( i \) by the first rater and category \( j \) by the second rater. Similarly, \( E_{ij} \) denotes the expected frequency of agreement by chance alone, computed from marginal totals assuming independence. The quadratic weighting factor \( w_{ij} \) penalizes larger discrepancies between raters more heavily than smaller ones, with \( k \) being the total number of rating categories.

\subsection{JURE Configurations and Baselines} 
\label{sec:baselines}

We opt for a strong reasoning model, OpenAI's o1, as the Orchestrator. 
We then compare JURE-o1 against two strong baselines. Apart from the human annotators, we consider OpenAI's o1 itself as another strong baseline. In this way, any performance gain of JURE can be immediately attributed to the addition of our Expert Pool and framework design.

\section{Results and Analysis}
\label{sec:results-analysis}

\subsection{Inter-Evaluator Agreement}

\begin{wrapfigure}{r}{0.48\textwidth}
    \vspace{-13pt}
    \centering
    \begin{minipage}{\linewidth}
        \includegraphics[width=\linewidth]{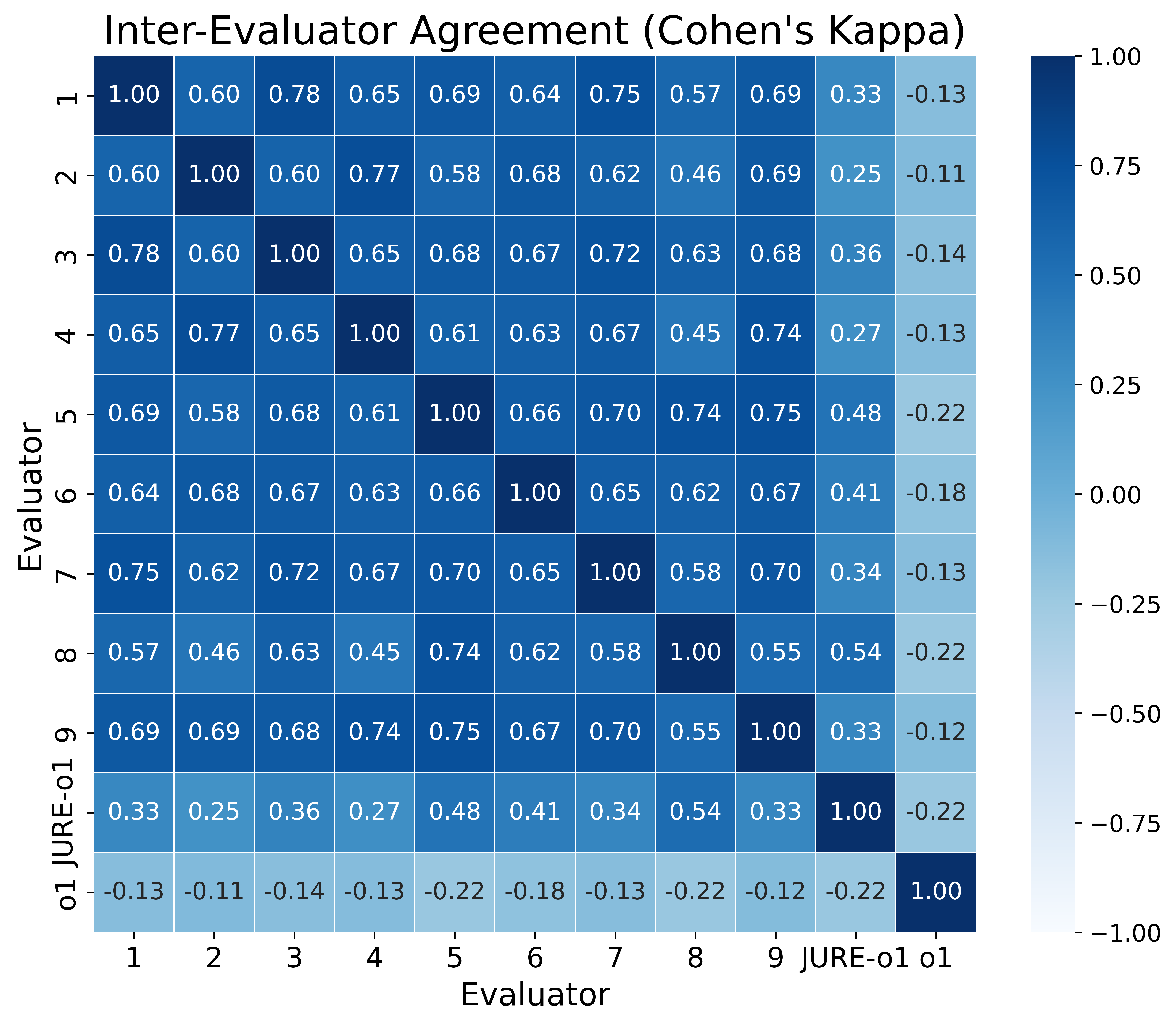}
        \caption{Inter-Evaluator Agreement (Cohen’s Kappa) Heatmap. Values represent agreement levels between evaluators, where higher values indicate stronger alignment.}
        \label{fig:inter-evaluator-heatmap}
    \end{minipage}
    \vspace{-40pt}
\end{wrapfigure}

To assess the consistency of evaluation across different annotators and models, we compute Cohen's Kappa correlation among the nine human annotators, o1, and JURE-o1. The results are visualized in Figure~\ref{fig:inter-evaluator-heatmap}, which presents a heatmap of agreement scores. Key findings are listed below.

\paragraph{Human-Human Agreement.}
Human annotators exhibit considerable variability in their agreement, with Cohen's Kappa scores ranging from approximately $0.45$ to $0.78$. This variation reflects the inherent subjectivity and complexity of evaluating IIE results. Given this diversity in human judgment, an evaluator achieving a correlation within this range can thus be regarded as reaching human-level judgment quality.

\paragraph{o1 vs. Human Annotators.}
The agreement between o1 and human annotators is notably poor, with Cohen’s Kappa values all falling below zero, ranging from $-0.22$ to $-0.12$. This performance is significantly lower than the human-human agreement baseline, underscoring the substantial gap between o1’s evaluations and human judgment. Despite being a powerful MLLM with advanced reasoning capabilities, o1 struggles to reliably assess intricate IIE tasks, highlighting the limitations of a single-model evaluation approach.

\paragraph{JURE-o1 vs. Human Annotators.}  
JURE-o1 consistently achieves higher correlation scores, with Cohen's Kappa value $0.54$ matching human-human agreement ($\geq 0.45$). This indicates that by leveraging dynamic routing to the right expertise, JURE-o1 attains human-level judgment quality, effectively closing the performance gap observed in o1. Notably, this improvement is entirely attributable to the expert routing mechanism, highlighting its effectiveness.

\subsection{Expert Routing Analysis}

Overall, the JURE framework demonstrates an optimal expert allocation, ensuring that each sub-task is evaluated with the most relevant expertise.

\begin{wrapfigure}{r}{0.66\textwidth}
    \vspace{-0pt}
    \centering
    \begin{minipage}{\linewidth}
        \includegraphics[width=\linewidth]{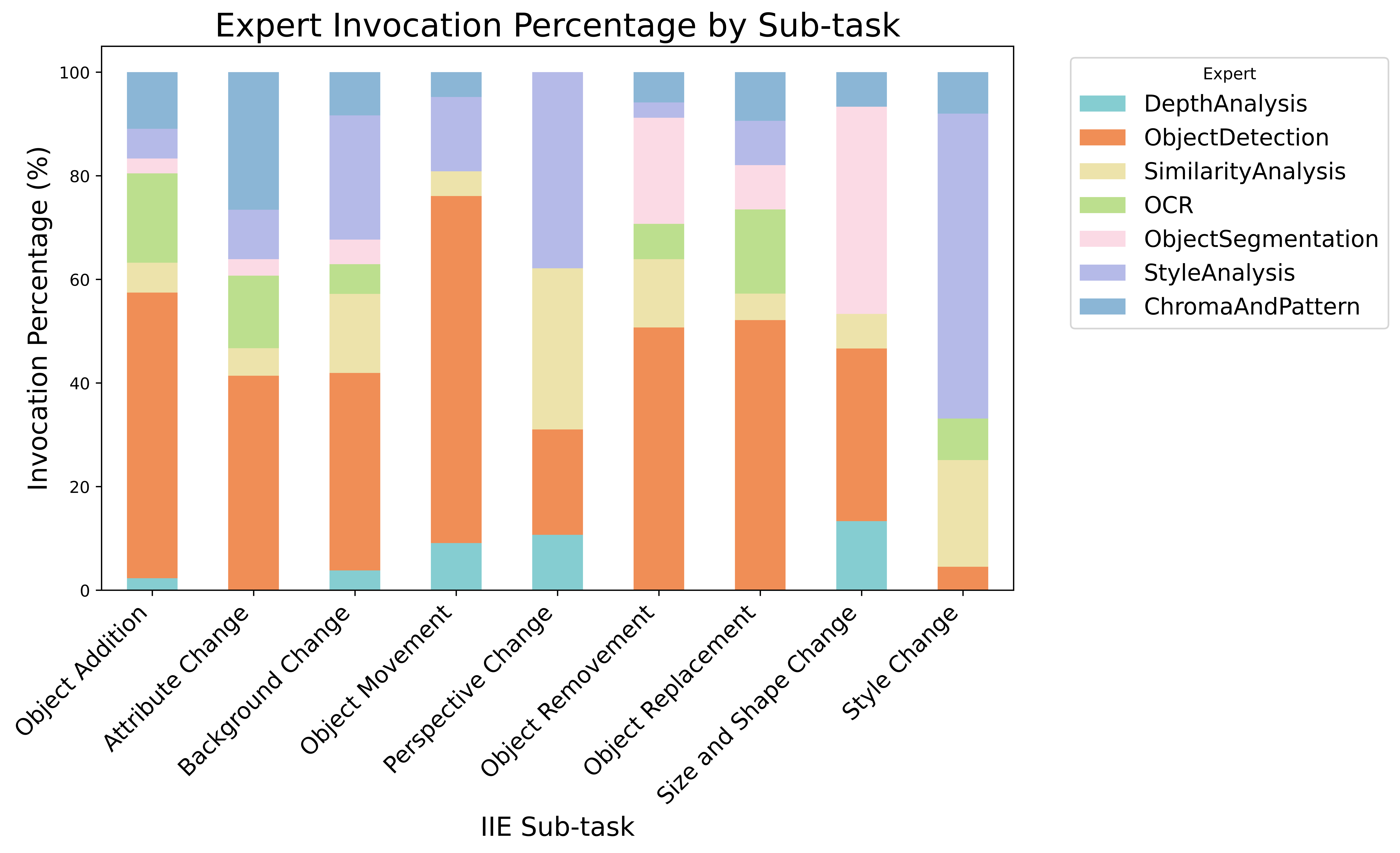}
        \caption{Expert Invocation Frequency in JURE-o1. 
        }
        \label{fig:inter-evaluator-heatmap}
    \end{minipage}
    \vspace{-10pt}
\end{wrapfigure}

Object Detection remains the most frequently invoked expert across most tasks, particularly for object-based edits such as Object Addition (55.17\%), Replacement (52.13\%), Movement (67\%) and Removement (50.73\%). We also manually inspect some cases from subtasks like Background Change and Perspective Change that are less relevant to direct object manipulations, JURE also naturally checks the existence of the objects in the edited images to ensure no object is missing, reflecting its central role across various IIE tasks.

Similarity Analysis is the second import expert accordingly the experiment results. Unlike other experts that usually takes one image as the input at each invocation, similarity analysis can take the original image along with one or more edited images as the input and return the similarity analysis compared to the original image. With manual inspections of the routing trajectories, we can confirm that it almost gets invoked for every evaluation. It takes a large portion in Background Change (15.26\%), Perspective Change (31.09\%) and Style Change (20.6\%), where the similarity between the edited images and the original one should definitely be analyzed in detail.

Style Analysis plays a dominant role in Style Change (58.87\%) and is routed appropriately for tasks like Background Change (23.98\%) and Perspective Change (37.87\%), ensuring that styles are accurately changed or maintained. Chroma and Pattern further enhances the evaluation of visual coherence, particularly in tasks like Attribute Change (10.92\%), where color and texture modifications are critical.

Depth Analysis is routed effectively for tasks involving spatial changes such as Object Movement (9.09\%), Perspective Change (10.70\%) and Size and Shape Change, ensuring JURE captures depth-related adjustments that affect the spatial layout of the scene.

Another interesting finding is that JURE can meaningfully chain multiple experts together in a non-trivial, task-aware manner.
As shown in Figure~\ref{fig:jure-architecture}, it can creatively feed bounding boxes from Dino-X to SAM2, rather than directly prompt SAM2 with textual inputs, to get a more precise segmentation masks. The Orchestrator further use these masks to crop the edited regions across all candidate outputs. It subsequently invokes the Similarity Analysis expert on the masked-out regions, ensuring that the analysis focuses solely on the areas that should not be modified. Similar chaining actions consistently happens across other experts.

\section{Conclusion}
\label{sec:discussion}
In this work, we addressed the critical challenge of evaluating IIE outputs with human-aligned judgment. JURE is a modular, explainable and future-proof evaluation framework that dynamically routes evaluation tasks to specialized experts based on the specific editing instruction and the feedback from each expert invocation.

For future work, we suggest: \emph{(1)} Fine-tuning the Orchestrator and experts with in-domain data to enhance task-specific performance. \emph{(2)} Exploring end-to-end optimization via reinforcement learning. Given only human preference scores (without explanations), the Orchestrator and experts can be jointly trained to achieve human-level, explainable judgment. \emph{(3)} Extending JURE beyond IIE evaluation, not only as a general-purpose judging framework, but also as a modular architecture for building general intelligent systems—offering a new paradigm centered on explicit routing and composable expertise.

\bibliography{colm2025_conference}
\bibliographystyle{colm2025_conference}

\newpage
\appendix
\section{Algorithm Workflow}
\label{appendix:algorithm}

This section provides a detailed description of the JURE framework's workflow, highlighting how the orchestrator dynamically routes tasks to specialized experts during the evaluation process. Algorithm~\ref{alg:jure-main} outlines the structured steps of this iterative process, ensuring that each evaluation dimension is accurately assessed through expert involvement.

Initially, the orchestrator initializes the context and available experts. It then enters a loop, continuously invoking the \texttt{OrchestratorReasoning} function to determine which expert to engage next based on the current evaluation state stored in \texttt{Context}. Experts are called dynamically, with their outputs incrementally refining and adjusting the evaluation trajectory until no further expertise is needed. The final aggregated results are then returned as a comprehensive evaluation report and scores.

\begin{algorithm}[H]
\footnotesize
\caption{JURE Workflow}
\label{alg:jure-main}
\begin{algorithmic}
\Require 
    \textit{EditingInstruction}, \textit{Images}, \textit{ExpertPool}
\Ensure 
    \textit{FinalReport}

\\
\State \textit{// Orchestrator and the main evaluation loop}
\Function{RunJURE}{Inputs}
    \State \texttt{Context} $\gets$ Initialize with input data
    \State \Call{InitializeExperts}{ExpertPool}
    \\
    \While{True}
        \State \texttt{Expert, Args} $\gets$ \Call{OrchestratorReasoning}{Context, ExpertPool}
        \If{\texttt{Expert} = None}
            \State \texttt{FinalReport} $\gets$ \Call{AggregateResults}{Context}
            \State \Return \texttt{FinalReport}
        \EndIf
        \State \texttt{ExpertOutput} $\gets$ \Call{InvokeExpert}{Expert, Args}
        \State \Call{UpdateContext}{Context, ExpertOutput}
    \EndWhile
    \\
\EndFunction

\\
\Function{OrchestratorReasoning}{Context, ExpertPool}
    \State Analyze \texttt{Context} to determine evaluation gaps
    \If{No additional evaluations needed}
        \State \Return None, None
    \Else
        \State Select \texttt{Expert} and prepare \texttt{Args} based on \texttt{Context}
        \State \Return \texttt{Expert}, \texttt{Args}
    \EndIf
\EndFunction

\\
\State \textit{// Expert Pool: Example experts}

\Function{ObjectDetectionExpert}{Image}
    \State Instantiated with DINO-X
    \State \texttt{BoundingBoxes} $\gets$ Analyze \texttt{Image}
    \State \Return \texttt{BoundingBoxes}
\EndFunction

\\
\dots

\end{algorithmic}
\end{algorithm}

\section{IIE Sub-tasks and Examples}
\label{appendix:iie-subtasks}

This section elaborates on the sub-tasks involved in Instruction-based Image Editing (IIE), providing clarity through concrete examples. Table~\ref{appendix:tab:iie-subtasks} summarizes representative editing tasks along with example instructions and their distribution in the Emu test set.

Each listed sub-task illustrates typical instructions users might provide to an IIE model, ranging from simple edits such as \textit{Object Removal} ("Remove the lamp from the table.") to complex edits like \textit{Perspective Editing} ("Set the camera to a bird’s-eye view."). 

\begin{table}[h!]
\centering
\caption{Representative sub-tasks in IIE along with Emu testset distribution.}
\label{appendix:tab:iie-subtasks}
\begin{tabular}{l p{0.45\textwidth} r}
\toprule
\textbf{Sub-task} & \textbf{Example Instruction} & \textbf{\#Instances (\%)} \\
\midrule
Object Addition  & ``Add a bouquet of flowers on the table.'' & 550 (28.59\%) \\
Object Replacement  & ``Replace the cat with a teddy bear.'' & 204 (10.60\%) \\
Object Movement  & ``Move the chair closer to the window.'' & 13 (0.68\%) \\
Object Removal  & ``Remove the lamp from the table.'' & 285 (14.81\%) \\
Background Change  & ``Change the background to a beach.'' & 391 (20.32\%) \\
Attribute Change  & ``Change the car’s color to metallic red.'' & 322 (16.74\%) \\
Style Change  & ``Render the scene in a oil painting style.'' & 241 (12.53\%) \\
Size/Shape Modification  & ``Resize the tree to make it taller.'' & 6 (0.31\%) \\
Perspective Editing  & ``Set the camera to a bird’s-eye view.'' & 4 (0.21\%) \\
\bottomrule
\end{tabular}
\vspace{-15pt}
\end{table}

\end{document}